\documentclass[letterpaper, 10 pt, conference]{ieeeconf}  

\IEEEoverridecommandlockouts                              

\RequirePackage{xcolor, hyperref} 
\usepackage{cite}
\usepackage[normalem]{ulem}

\usepackage{graphicx}
\usepackage{epsfig} 
\usepackage{textcomp}
\usepackage{pifont}

\usepackage{amsmath,amssymb,amsfonts}
\usepackage{algorithm}
\usepackage{algpseudocode}
\usepackage{multirow}
\usepackage{booktabs}
\usepackage[flushleft]{threeparttable}

\usepackage{xcolor}
\usepackage{colortbl}
\usepackage{hyperref}
\hypersetup{
  colorlinks=true,         
  linkcolor=blue, 
  citecolor=blue,
  filecolor=magenta,
  urlcolor=black  
}

\newcommand{\cc}[1]{\cellcolor[HTML]{EFEFEF}}
\newcommand{\rc}[1]{\rowcolor[HTML]{EFEFEF}}
\algrenewcommand\algorithmicrequire{\textbf{Input:}}
\algrenewcommand\algorithmicensure{\textbf{Output:}}
\algrenewcommand\algorithmiccomment[1]{// {\itshape #1}}

\definecolor{subsectioncolor}{rgb}{0.0, 0.5, 0.5} 

\newcommand{\customref}[3][]{%
  \text{#2}~%
  \ifx\\#1\\%
    \hyperref[#3]{\ref*{#3}}%
  \else%
    \hyperref[#3]{\ref*{#3}(#1)}%
  \fi
}

\useunder{\uline}{\ul}{}

\title{\LARGE \bf
Aligning Humans and Robots via Reinforcement Learning from Implicit Human Feedback}

\author{
Suzie Kim$^{1}$, Hye-Bin Shin$^{2}$, and Seong-Whan Lee$^{1}$ 
\thanks{*This research was supported by the Institute of Information \& Communications Technology Planning \& Evaluation (IITP) grant, funded by the Korea government (MSIT) (No. RS-2019 II190079 (Artificial Intelligence Graduate School Program (Korea University)), and No. RS-2024-00457882 (AI Research Hub Project)).}%
\thanks{$^{1}$S. Kim and S.-W. Lee are with the Department of Artificial Intelligence, Korea University, Anam-dong, Seongbuk-ku, Seoul 02841, Korea. 
\tt\small \{sz\_kim, sw.lee\}@korea.ac.kr}%
\thanks{$^{2}$H.-B. Shin is with the Department of Brain and Cognitive Engineering, Korea University, Anam-dong, Seongbuk-ku, Seoul 02841, Korea.
\tt\small hb\_shin@korea.ac.kr}%
}

\begin{document}

\maketitle
\thispagestyle{empty}
\pagestyle{empty}

\begin{abstract}

Conventional reinforcement learning (RL) approaches often struggle to learn effective policies under sparse reward conditions, necessitating the manual design of complex, task-specific reward functions. To address this limitation, reinforcement learning from human feedback (RLHF) has emerged as a promising strategy that complements hand-crafted rewards with human-derived evaluation signals. However, most existing RLHF methods depend on explicit feedback mechanisms such as button presses or preference labels, which disrupt the natural interaction process and impose a substantial cognitive load on the user.
We propose a novel reinforcement learning from implicit human feedback (RLIHF) framework that utilizes non-invasive electroencephalography (EEG) signals, specifically error-related potentials (ErrPs), to provide continuous, implicit feedback without requiring explicit user intervention. The proposed method adopts a pre-trained decoder to transform raw EEG signals into probabilistic reward components, enabling effective policy learning even in the presence of sparse external rewards.
We evaluate our approach in a simulation environment built on the MuJoCo physics engine, using a Kinova Gen2 robotic arm to perform a complex pick-and-place task that requires avoiding obstacles while manipulating target objects. The results show that agents trained with decoded EEG feedback achieve performance comparable to those trained with dense, manually designed rewards. These findings validate the potential of using implicit neural feedback for scalable and human-aligned reinforcement learning in interactive robotics.

\end{abstract}


\section{INTRODUCTION}\label{sec:intro}

Reinforcement learning (RL) has emerged as a promising paradigm for training agents to perform complex robotic tasks such as manipulation and locomotion. However, achieving competent task execution often hinges on the design of handcrafted, dense reward functions.
Designing such reward signals is costly, requiring extensive manual tuning and domain-expert knowledge. Furthermore, these functions are typically task-specific and lack generalizability, thereby limiting the scalability and broader applicability of conventional RL methods.

To address these limitations, reinforcement learning from human feedback (RLHF)~\cite{rlhf} has emerged as a promising alternative. Rather than depending on explicitly specified reward functions, RLHF leverages subjective human evaluations as learning signals, enabling policy optimization without the need for intricate reward engineering~\cite{leeCurling}. Nevertheless, existing RLHF approaches typically depend on explicit feedback mechanisms, such as button presses, trajectory annotations, or preference comparisons~\cite{pebble, surf, rune, rime, breadcrumbs}, which impose substantial cognitive load and interrupt the natural flow of interaction. Therefore, the development of practical RLHF systems necessitates the integration of intrinsic human feedback that is minimally intrusive and capable of delivering continuous signals throughout interaction.

\begin{figure}[!t]
\centerline{\includegraphics[width=0.95\columnwidth]{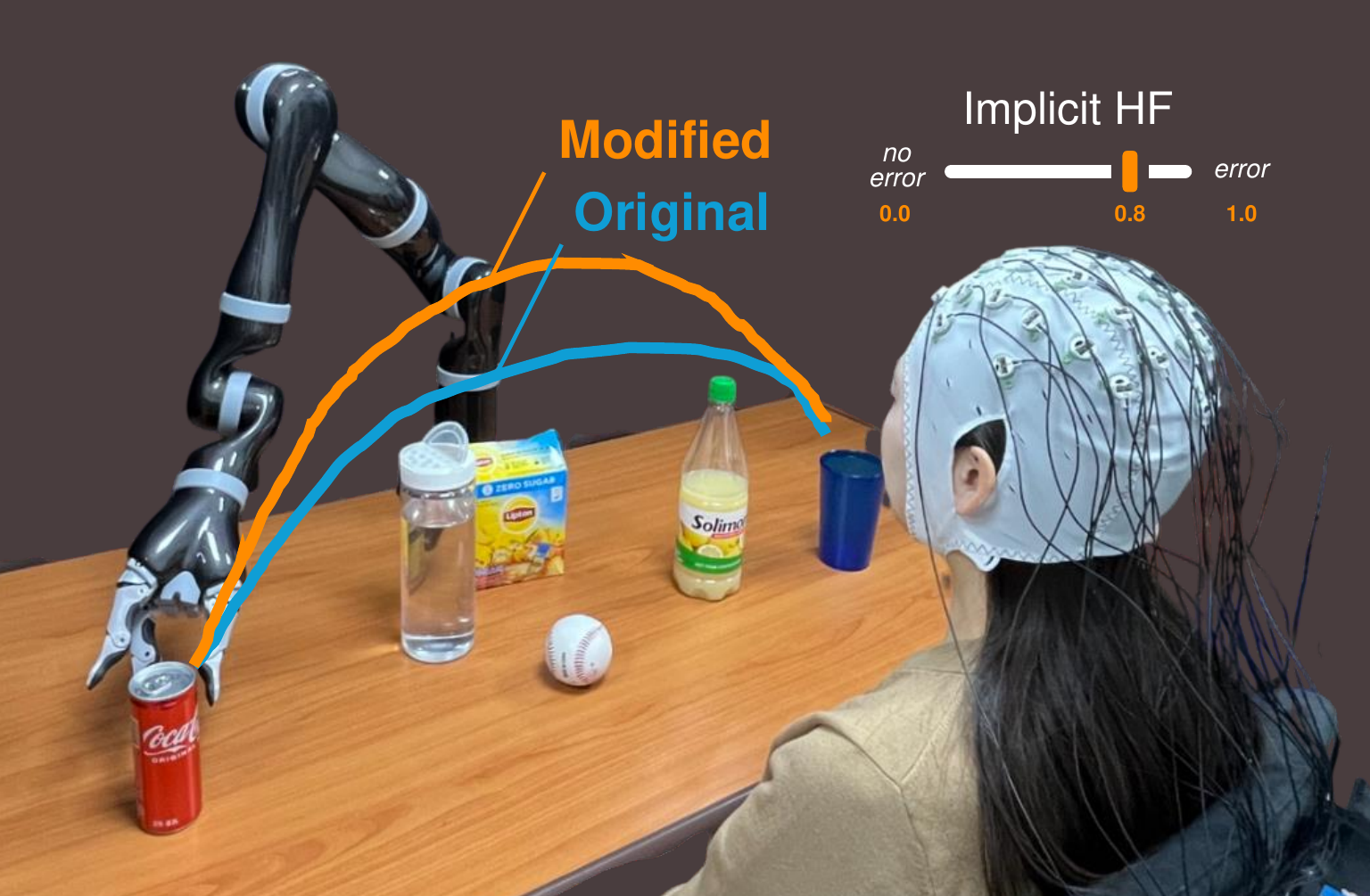}}
\caption{Conceptual illustration of the proposed RLIHF framework in a real-world pick-and-place scenario. A human user wearing an EEG cap observes a robotic arm navigating a cluttered tabletop environment. The “Original” trajectory (blue) minimizes path length but approaches obstacles too closely, violating implicit spatial preferences. In contrast, the “Modified” trajectory (orange), guided by EEG-based human feedback, maintains safer clearances. EEG signals are decoded to estimate the probability of perceived error, which is transformed into a continuous reward signal used to adapt the behavior of the robot.}
\vspace{-0.5cm}
\label{fig:fig1}
\end{figure}

In this work, we propose a novel Reinforcement Learning from Implicit Human Feedback (RLIHF) framework that enables policy learning based on real-time feedback derived from neural signals, without requiring explicit human intervention. The proposed framework utilizes non-invasive electroencephalography (EEG) signals to continuously reflect the human observer’s internal evaluation of robotic behavior and adapt the agent’s policy accordingly. Specifically, we leverage error-related potentials (ErrPs), stereotypical EEG responses spontaneously elicited when a human detects erroneous robot actions. This allows the agent to receive semantically meaningful feedback without the manual reward design or labeling. Whereas prior studies have predominantly treated ErrPs as discrete events~\cite{hri, accelerating}, our framework continuously decodes the probability of error occurrence and integrates this signal into the reward function. This design enables agents to progressively refine their policies even in environments where external rewards are sparse or delayed.

To decode ErrPs from EEG input, we employ a neural classifier based on EEGNet~\cite{eegnet}, a lightweight convolutional architecture pre-trained on a pooled dataset of labeled EEG signals collected from multiple subjects. The decoder remains fixed during training to ensure stability and signal consistency, but can be infrequently updated online to accommodate changes in individual neural response patterns. Although decoder accuracy varies across participants, with some individuals showing only marginally above-chance classification performance, the proposed RLIHF framework demonstrates that even imperfect decoders can produce reward signals that are sufficiently informative for effective policy learning. Notably, agents trained under RLIHF achieve performance comparable to those trained with fully engineered dense rewards, despite the variability in decoder quality. This robustness to decoder performance heterogeneity constitutes an additional contribution of our method.

\begin{figure}[!t]
\centerline{\includegraphics[width=\columnwidth]{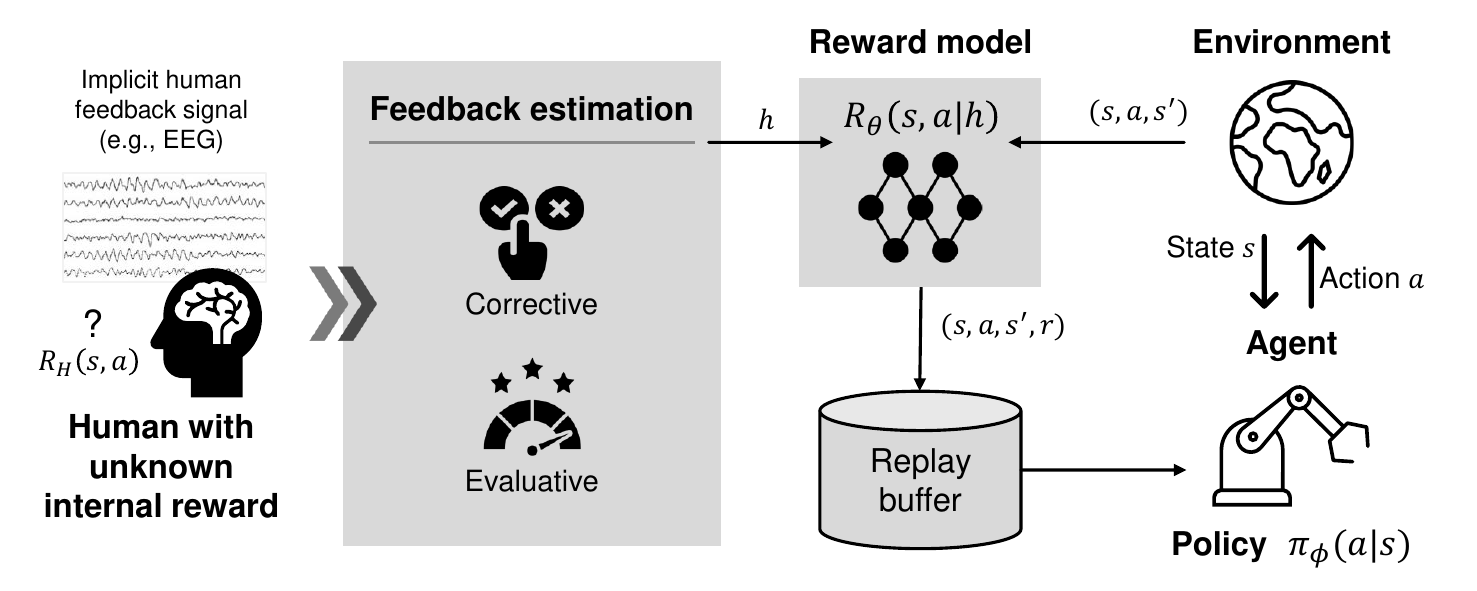}}
\caption{Overview of the proposed RLIHF framework. Implicit human feedback is decoded into scalar rewards and integrated with environment rewards. The resulting rewards guide policy updates via Soft Actor-Critic, with transitions stored in a replay buffer for sample-efficient learning.}
\vspace{-0.5cm}
\label{fig:fig2}
\end{figure}

The RLIHF framework mitigates the limitations of sparse reward settings by transforming automatically acquired neural responses into dense and informative reward signals. To validate these contributions, we conducted experiments in a MuJoCo-based \texttt{robosuite} simulation environment~\cite{robosuite}, where a Kinova Gen2 robotic arm was tasked with performing obstacle-avoiding pick-and-place operations. We compared three learning conditions—sparse reward, dense reward, and RLIHF—by training a policy under each condition and evaluating them in an independent test environment. The results demonstrate that RLIHF significantly outperforms the sparse condition and achieves performance comparable to dense reward setups. By directly incorporating decoded neural signals into the training process, our method supports human-aligned policy refinement without requiring handcrafted supervision. These characteristics make it well suited not only for general human–robot interaction (HRI) tasks, but also for brain–computer interface (BCI) systems, where implicit feedback must be processed in real time. Potential applications include collaborative robotics, fine-grained teleoperation, and personalized assistive technologies, where continuous adaptation to user preferences is essential.

\section{RELATED WORKS}\label{sec:relatedworks}

\subsection{Reinforcement Learning from Human Feedback}

In real-world robotic applications, RL often suffers from sparse or delayed rewards, which severely hinder exploration and make policy convergence unstable~\cite{accelerated}. Classic approaches such as reward shaping~\cite{prefT}, inverse reinforcement learning (IRL)~\cite{customizing}, and behavioral cloning have been proposed to mitigate this bottleneck by incorporating domain knowledge or mimicking expert behavior. While these methods improve sample efficiency, they often require precise reward engineering, labor-intensive demonstrations, or manually designed feature representations, which limit their adaptability across tasks.


RLHF has emerged as a more flexible alternative, enabling agents to optimize policies using subjective human preferences rather than explicit scalar rewards~\cite{rlhf}. Methods such as preference ranking~\cite{prefT}, trajectory comparison~\cite{surf, rune}, and interaction-based scoring~\cite{interactive, deeptamer} have demonstrated the ability to convey nuanced task objectives that are difficult to formalize analytically. However, most RLHF frameworks rely heavily on explicit feedback modalities—button presses~\cite{pebble}, trajectory annotations~\cite{latte}, or comparative labels~\cite{breadcrumbs}—which impose substantial cognitive demands and disrupt the natural flow of interaction. These characteristics make them difficult to deploy in high-frequency or real-time human-in-the-loop scenarios, motivating the exploration of implicit feedback channels.

\subsection{Leveraging EEG Signals as Human Feedback}

Implicit feedback modalities have recently gained traction as a promising direction for reducing human supervision burden in RL~\cite{affect, empathic}. Among these, non-invasive neural signals captured via EEG provide a continuous, low-latency window into users’ internal states without requiring overt responses~\cite{hitl}. In particular, ErrPs—a class of event-related potentials evoked when a human perceives an error—have been shown to correlate with evaluative judgments of agent behavior~\cite{accelerated, maximizing}. ErrPs are attractive as a reward proxy because they are automatically and involuntarily generated, allowing evaluative feedback to be harvested without interrupting task execution.


Prior work has explored using ErrPs for policy correction in RL, often treating them as binary error detection triggers or sparse intervention signals~\cite{delpreto, singletrial, leeSchizophrenia, leeNeuroGrasp, leeAbstract, leeNew}. These methods typically operate by identifying high-confidence error events and adjusting the agent’s behavior retrospectively, such as by relabeling trajectories~\cite{customizing} or issuing corrective control commands~\cite{accelerating}. While effective in simple tasks, this binary interpretation fails to capture the nuanced and probabilistic nature of neural feedback, thereby underutilizing the information content embedded in EEG signals. Moreover, many of these systems require online decoder adaptation or per-user calibration, which compromises their scalability in practical settings.

\section{MATERIALS AND METHODS}\label{sec:method}

\subsection{Proposed Framework}

We propose a novel RLIHF framework that integrates the brain-derived evaluative feedback into reward shaping for robotic policy learning. This framework enables adaptation based on internal human evaluations without requiring explicit interventions. As illustrated in Fig.~\ref{fig:fig2}, we leverage ErrPs, EEG responses to perceived errors, offer reliable implicit feedback due to their temporal resolution and consistent elicitation. Although the EEG data were replayed from an offline dataset~\cite{hri}, we used a streaming ring buffer to feed time-aligned epochs to the classifier, preserving real-time dynamics and ensuring compatibility with closed-loop deployment. These neural signals are decoded in real-time and mapped to scalar rewards, allowing the agent to adjust its policy based on perceived internal evaluative signals.

To extract ErrPs from EEG data, we employ EEGNet~\cite{eegnet}, a compact convolutional neural network tailored for EEG-based BCI applications. Given an input EEG epoch $\mathbf{x} \in \mathbb{R}^{C \times T}$, where $C$ denotes the number of channels and $T$ the number of sampled time points, the model predicts a class distribution:
\begin{equation}
    \mathbf{p} = \text{softmax}(f_{\theta}(\mathbf{x})),
\end{equation}
where $f_\theta$ is the EEGNet model and $\mathbf{p}_1$ corresponds to the predicted probability of an ErrP. The classifier is trained using cross-entropy loss on labeled EEG trials:
\begin{equation}
    \theta_{\text{ErrP}}^* = \arg\min_{\theta_{\text{ErrP}}} \frac{1}{N} \sum_{i=1}^{N} L_{\text{ErrP}}(f(x_i; \theta_{\text{ErrP}}), e_i),
    \label{eq:eeg_training}
\end{equation}
where $x_i$ is an EEG segment, $e_i$ is the binary error label, and $L_{\text{ErrP}}$ is the loss function. During execution, the classifier receives preprocessed EEG segments and outputs the estimated likelihood of error:
\begin{equation}
    p_{\text{ErrP}} = \mathbf{p}_1,
\end{equation}
which is transformed into a scalar reward:
\begin{equation}
    r_{t}^{\text{ErrP}} = 1 - p_{\text{ErrP}}.
\end{equation}
This decoded reward serves as a continuous form of implicit human feedback, directly integrated into the RL update rule to guide policy optimization within the RLIHF framework. To improve learning stability and align the reward signal with task-level objectives, this neural reward is further combined with task-specific environmental signals, such as success events or obstacle collisions, to form a composite reward used during policy training.

We adopt the Soft Actor-Critic (SAC) algorithm~\cite{sac} for agent training due to its compatibility with the demands of EEG-based feedback. As an off-policy method, SAC stores transitions in a replay buffer, allowing efficient reuse of EEG-labeled interactions. SAC also supports continuous action spaces and stochastic policy learning, essential for fine-grained control. Its entropy-regularized objective encourages exploration under uncertainty, mitigating the risk of convergence to suboptimal policies in early training phases. Compared to on-policy algorithms such as Proximal Policy Optimization (PPO)~\cite{ppo}, which require frequent sampling, or value-based methods like Deep Q-learning (DQN)~\cite{dqn} that are restricted to discrete actions and sensitive to noise, SAC offers a balanced and robust framework. Its sample efficiency, robustness to uncertainty, tolerance to noisy feedback, and ability to handle continuous control tasks make it particularly well-suited for human-in-the-loop reinforcement learning based on implicit EEG signals.

\begin{figure}[!t]
\centerline{\includegraphics[width=0.95\columnwidth]{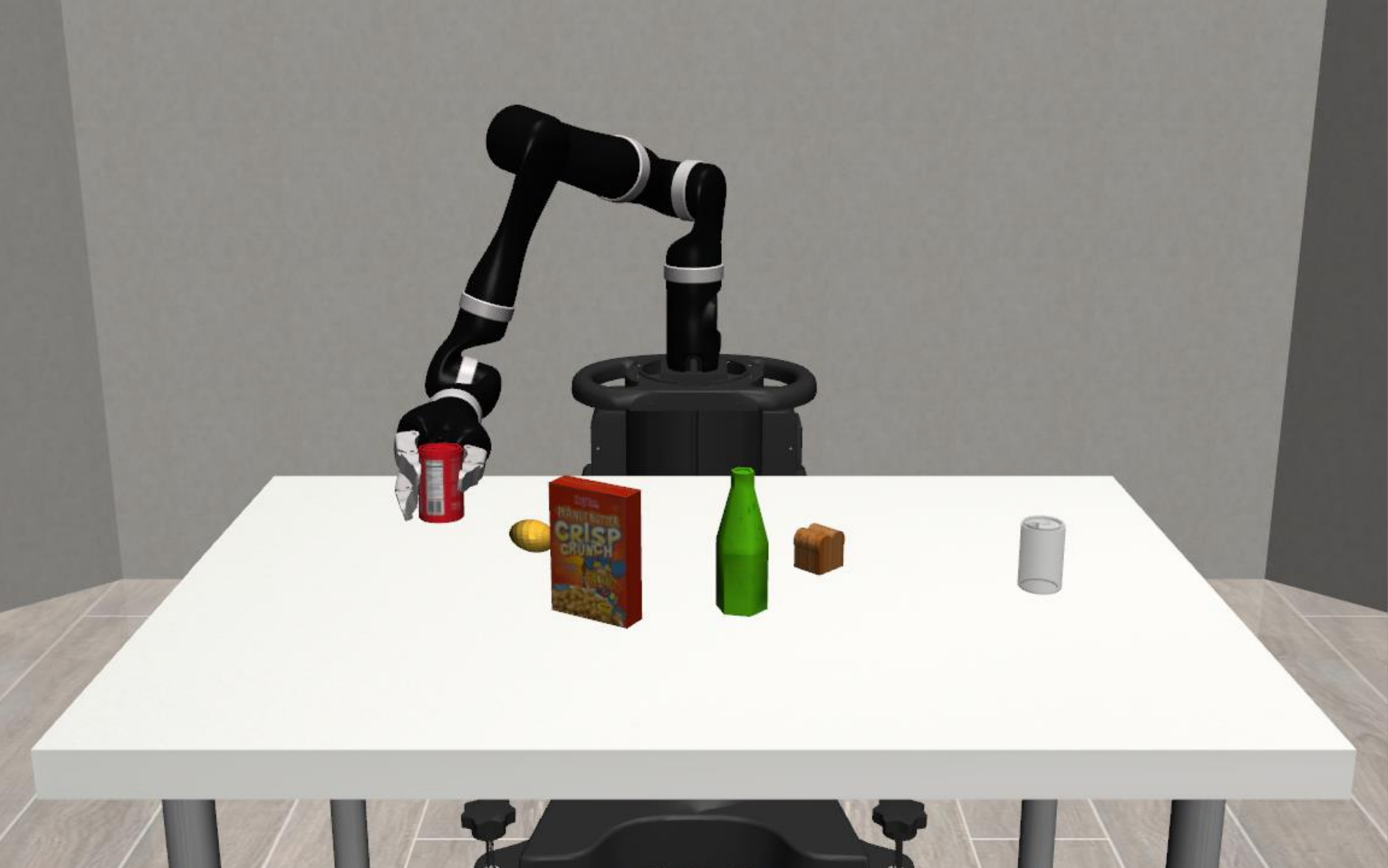}}
\caption{Customized pick-and-place simulation environment used for training and evaluation in the RLIHF framework. A Kinova Gen2 robotic arm operates in a cluttered workspace containing four obstacles (lemon, cereal box, green bottle, and bread). The task requires the robot to grasp the designated target object (red can) and place it at the goal location, marked by a gray cylinder on the right, while avoiding collisions. The environment was built by modifying the \texttt{Lift} task in the \texttt{robosuite} framework and executed within the MuJoCo physics engine.}
\vspace{-0.5cm}
\label{fig:fig3}
\end{figure}

\subsection{HRI Task}

The learning framework is evaluated in a simulated pick-and-place task using a Kinova Gen2 robotic arm operating in a cluttered tabletop workspace populated with everyday household objects. The environment is implemented with the MuJoCo physics engine and structured through the \texttt{robosuite} interface. This setup introduces non-trivial navigational challenges that go beyond basic motion planning, requiring the agent to reason about more than just collision-free trajectories.

The agent must optimize task efficiency while implicitly adhering to spatial constraints, such as maintaining safe distances from surrounding objects. Rather than relying on hard-coded obstacle avoidance, the agent is required to infer and adapt to implicit spatial preferences that are difficult to capture through traditional reward design. Human users typically prefer trajectories that maintain a margin of clearance from nearby obstacles, even when those paths are longer or less time-efficient. The central difficulty is in learning policies that reflect such nuanced human expectations, where comfort, clearance, and control must all be considered simultaneously, under conditions of scarce explicit supervision and limited access to manually engineered dense rewards. ErrP feedback offers a cognitively unobtrusive yet semantically meaningful signal that allows the agent to infer such preferences and adjust its behavior to align with implicit human evaluations throughout training.

\begin{figure*}[!ht]
\centerline{\includegraphics[width=\textwidth]{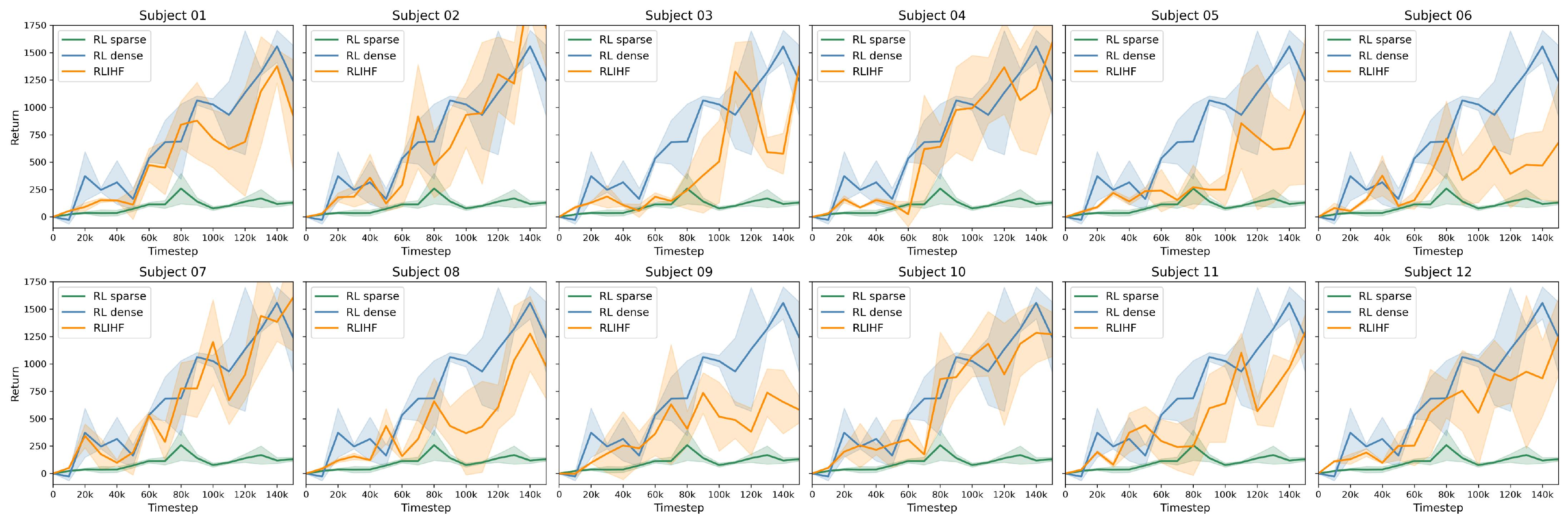}}
\caption{
Evaluation performance curves for the RL baselines (sparse and dense), and our proposed RLIHF method across 12 human subjects.
Each plot shows the mean episodic return during evaluation, averaged across five independent runs. The RLIHF agent consistently outperforms the sparse reward baseline and often approaches the ideal performance achieved with dense rewards, with some inter-subject variability. 
}
\vspace{-0.5cm}
\label{fig:fig4}
\end{figure*}

\subsection{HRI-ErrP Dataset}

To train the EEG classifier, we utilized a publicly available EEG dataset collected from 12 participants in a real-world HRI setting, where ErrPs were elicited as participants observed robotic actions that violated expected behavior\cite{hri}. The experimental protocol was specifically designed to capture implicit neural responses to perceived robot errors, providing a reliable source of feedback grounded in real-time human evaluation.

EEG signals were recorded using a 32-channel actiChamp system at 1000 Hz and downsampled to 256 Hz. We applied bandpass filtering between 1--20 Hz and re-referencing, and segmented the data into 2-second windows aligned to the feedback onset. Each segment was labeled as either an error or non-error event depending on whether the corresponding trial involved an observed robot error. The resulting dataset was used to train an EEGNet-based classifier in a leave-one-subject-out (LOSO) cross-validation setting. The trained model, along with the preprocessed EEG dataset, was subsequently used to simulate streaming human feedback for policy training in our RLIHF framework.

\subsection{Performance Evaluation}

We compare our RLIHF method against two baseline conditions. The first is a sparse-reward baseline, in which the agent receives rewards only for successful task completion and is penalized for collisions between the gripper and surrounding obstacles. The second is a dense-reward baseline, which augments the sparse setting with additional engineered rewards based on how closely the agent’s trajectory aligns with an optimal path. To ensure comparability, all agents trained under sparse, dense, and RLIHF reward conditions were evaluated using a unified reward structure that accounted for task success, obstacle collisions, and trajectory alignment with a predefined expert path.

Training and evaluation were conducted as follows. Each episode consisted of 1000 timesteps, and each policy was trained for 150 episodes (150,000 timesteps). After training, we evaluated each policy at regular intervals by performing five rollouts per evaluation point and assessed performance in terms of mean return, return standard deviation, success rate, path efficiency, and path deviation. Success rate was defined as the proportion of episodes in which the agent successfully completed the pick-and-place task. Path efficiency measured the ratio between the length of the ideal trajectory and the agent's actual trajectory. Path deviation quantified the root mean squared deviation between the agent's executed trajectory and the ideal path. To ensure statistical robustness, each experiment was repeated using five different random seeds for each reward setting and participant. This standardized protocol enabled consistent and reliable comparison of policy performance across all experimental conditions.

\section{RESULTS AND DISCUSSION}\label{sec:results}

Fig.~\ref{fig:fig4} shows the evaluation return curves across 12 subjects under the sparse, dense, and RLIHF reward conditions. Across all subjects, learning with sparse rewards resulted in poor performance, indicating the challenge of learning without structured guidance. In contrast, RLIHF agents demonstrated significant improvements over sparse agents, achieving performance levels closely approaching those of dense reward agents. Examining the learning dynamics, we observed that during the Early training phase (0--50k timesteps), sparse, dense, and RLIHF agents exhibited relatively similar performance, reflecting the initial exploration phase. However, by the Mid phase (50k--100k timesteps), our method began to show a increase in return, narrowing the performance gap with dense agents. In the Late phase (100k--150k timesteps), the proposed agents achieved performance comparable to dense agents in several subjects, suggesting that implicit human feedback becomes increasingly influential as training progresses. This effect was also observed in subjects with weaker decoders, showing that our approach can still support learning under noisy feedback. This robustness underscores the strength of implicit signals even when supervision is limited. This phase-wise evolution highlights the potential for our framework to bootstrap learning even when initial classifier performance is moderate.

Table~\ref{tab:table} quantitatively compares the success rate, path efficiency, and path deviation over the course of training. 
RLIHF agents achieved significantly higher success rates than sparse agents across all phases, and exhibited success rates comparable to dense agents in the Mid and Late phases. Although path efficiency for RLIHF agents was slightly lower than for dense agents, this is expected: the task's objective was not to minimize path length but to align the robot's trajectory with the user's implicit preferences. This observation aligns with the design philosophy of our method, which emphasizes aligning agent behavior with human intent rather than achieving shortest-path efficiency. Importantly, path deviation values revealed that RLIHF agents maintained closer adherence to the ideal paths than sparse agents, confirming that EEG-derived feedback effectively steered robot behavior toward human-aligned trajectories. These results demonstrate that even under moderate decoding accuracy, implicit feedback can shape policy behavior toward interpretable and goal-directed motion.

\begin{table}[t]
    \centering
    \caption{Comparison of success rate, path efficiency, and path deviation across sparse, dense, and RLIHF reward settings at different training phases (Early, Mid, Late).}
    \begin{tabular}{l l c c c}
        \toprule
        \textbf{Phase} & \textbf{Method} & \textbf{Success Rate $\uparrow$} & \textbf{Path Eff. $\uparrow$} & \textbf{Path Dev. $\downarrow$} \\       
        \midrule
         & RL sparse & 0.00 ± 0.00 & 0.36 ± 0.20 & 0.74 ± 0.26 \\
        Early & RL dense & 0.06 ± 0.18 & 0.60 ± 0.31 & 0.50 ± 0.32 \\
         & RLIHF (ours) & 0.02 ± 0.12 & 0.59 ± 0.29 & 0.48 ± 0.20 \\
         \midrule
         & RL sparse & 0.01 ± 0.05 & 0.48 ± 0.28 & 0.53 ± 0.21 \\
        Mid & RL dense & 0.25 ± 0.41 & 0.71 ± 0.27 & 0.40 ± 0.14 \\
         & RLIHF (ours) & 0.17 ± 0.30 & 0.57 ± 0.23 & 0.45 ± 0.19 \\
         \midrule
         & RL sparse & 0.00 ± 0.00 & 0.60 ± 0.29 & 0.52 ± 0.17 \\
        Late & RL dense & 0.54 ± 0.45 & 0.74 ± 0.24 & 0.43 ± 0.07 \\
         & RLIHF (ours) & 0.39 ± 0.39 & 0.59 ± 0.21 & 0.45 ± 0.10 \\
        \bottomrule
    \end{tabular}
    \label{tab:table}
\end{table}

\begin{figure}[!t]
\centerline{\includegraphics[width=\columnwidth]{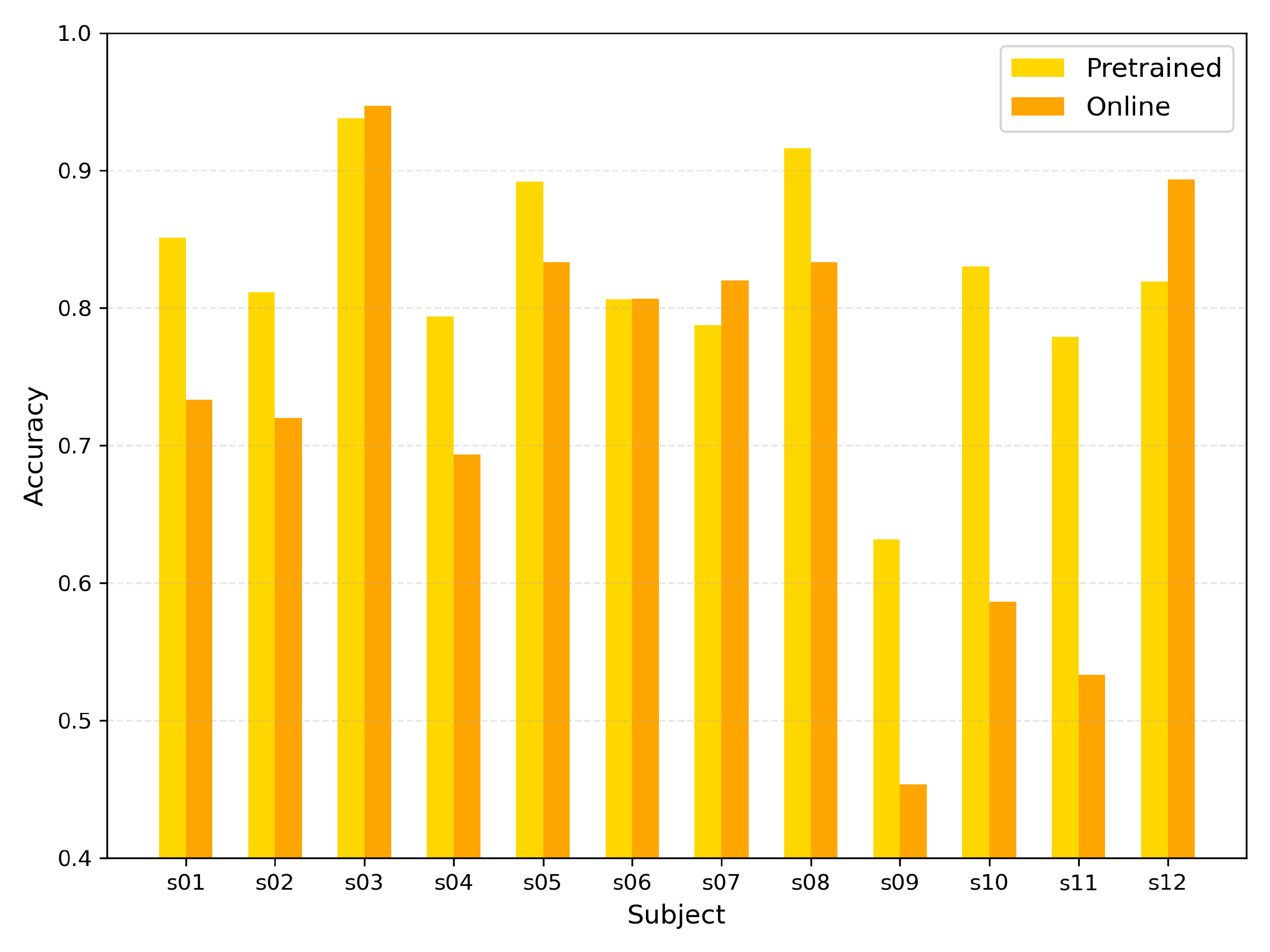}}
\caption{
ErrP decoding performance across 12 subjects. Yellow bars indicate accuracy achieved during pretraining, while orange bars represent online performance during real-time feedback integration. While higher decoding accuracy generally correlates with improved RLIHF return (see Fig.~\ref{fig:fig4}), we observe that performance above chance level is often sufficient to achieve comparable outcomes to dense reward learning.
}
\vspace{-0.5cm}
\label{fig:fig5}
\end{figure}

Fig.~\ref{fig:fig5} presents the pretraining and online classification accuracies of the ErrP classifier. Most subjects achieved classification accuracies well above chance level, typically ranging from 70\% to 90\%. Despite minor fluctuations between pretraining and online accuracies, policy performance remained robust, indicating that EEG classifiers with moderate but consistent accuracies are sufficient to guide effective policy learning. This suggests a degree of generalization capacity in the classifiers, enabling transfer from offline to online usage without catastrophic degradation. A subject-wise analysis revealed some variability across individuals. For example, in certain subjects, RLIHF agents even surpassed dense agents in terms of final returns, while in others, the gap remained slightly larger. This variability may stem from differences in EEG signal quality, the strength of individual ErrP responses, and subject-specific task engagement. Nevertheless, the overall trend remained consistent across the cohort, strengthening the generality of our findings.

\begin{figure}[!t]
\centerline{\includegraphics[width=\columnwidth]{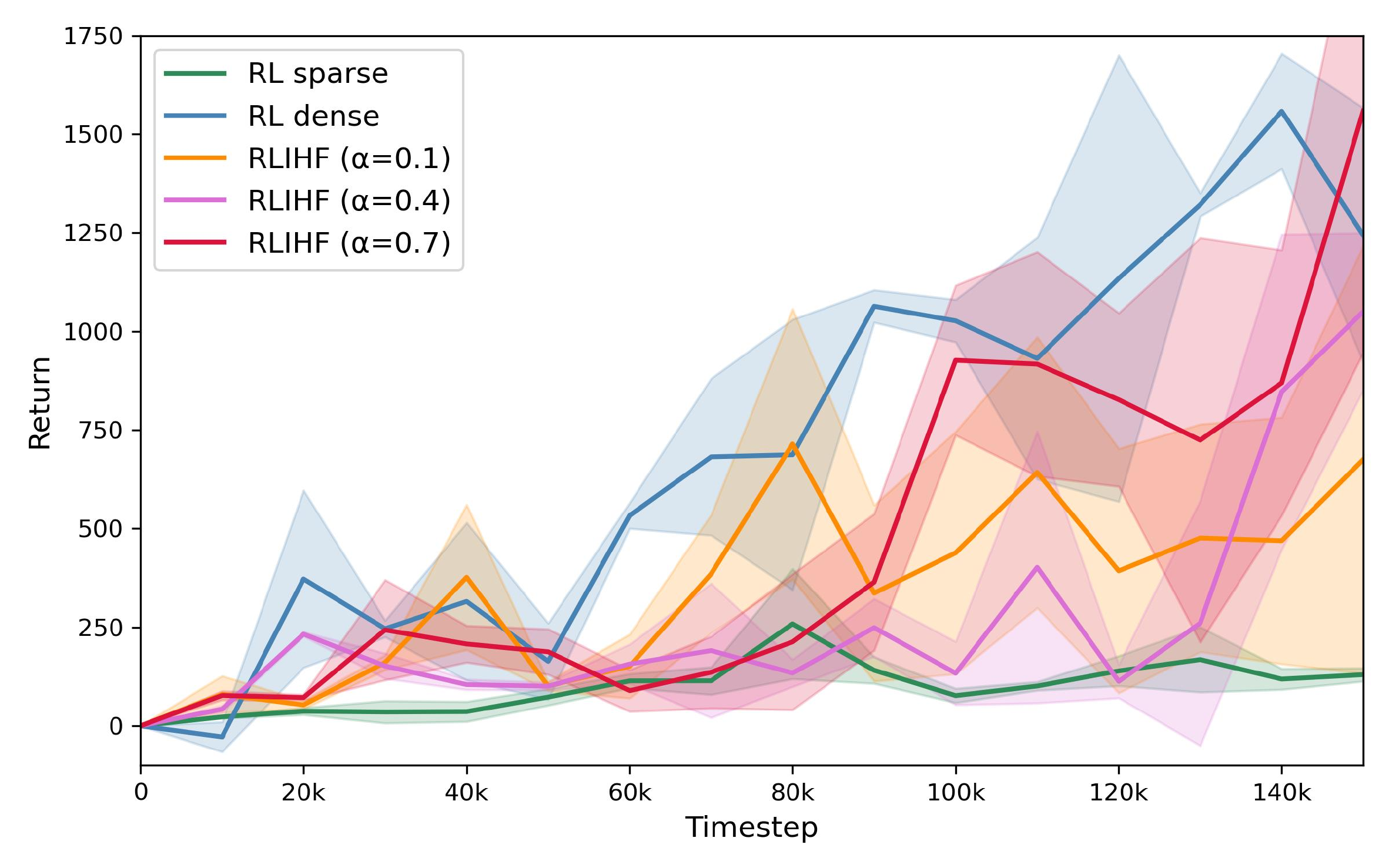}}
\caption{Effect of the human feedback weight parameter \( w_{\text{hf}} \) on RLIHF training performance. Episodic return curves are compared across three values of \( w_{\text{hf}} \) (\( \alpha=0.1, 0.4, 0.7 \)), alongside sparse and dense reward baselines. Increasing \( w_{\text{hf}} \) leads to greater reliance on EEG-derived human feedback, often resulting in more human-aligned behavior and improved Late-stage performance. The strong gain observed with \( \alpha = 0.7 \) suggests potential benefits of tuning feedback weighting in future RLIHF systems.}
\vspace{-0.5cm}
\label{fig:fig6}
\end{figure}

To further investigate the influence of the human feedback weight \(w_{\text{hf}}\), we conducted additional experiments by varying \(w_{\text{hf}}\) across three settings: \(w_{\text{hf}} = 0.1\) (default), \(w_{\text{hf}} = 0.4\), and \(w_{\text{hf}} = 0.7\) (see Fig.~\ref{fig:fig6}). The learning curves under each setting exhibited clear differences in adaptation dynamics. With \(w_{\text{hf}} = 0.1\), the incorporation of neural feedback was limited, and agents improved only gradually over time. At \(w_{\text{hf}} = 0.4\), moderate performance gains emerged, indicating partial utilization of the feedback signal. The \(w_{\text{hf}} = 0.7\) setting yielded both rapid learning progress and the most pronounced improvement in the Late stage of training, resulting in the highest overall returns. These results suggest that stronger weighting of human feedback can accelerate learning and ultimately enhance final performance, particularly when the feedback signal is sufficiently reliable.
Although not definitive, these observations suggest that adaptively modulating \(w_{\text{hf}}\) during training could be a promising direction for future work. One possible strategy is to begin with a lower weight to encourage exploration in the early phase, and then gradually increase it to emphasize human preferences as learning progresses.

The results consistently demonstrate the effectiveness of our RLIHF framework across a diverse subject pool. Compared to sparse rewards, RLIHF agents achieved significantly higher success rates and closer alignment with user-preferred trajectories, often approaching the performance of dense-reward agents. These trends were evident both in return curves and trajectory-level metrics, despite moderate ErrP decoding accuracy. Subject-wise variability did not undermine the general trend, confirming robustness to decoder differences. Additionally, our ablation study on feedback weighting showed that setting \(w_{\text{hf}} = 0.7\) led to the strongest learning outcomes, with the highest overall returns in the Late phase. By contrast, \(w_{\text{hf}} = 0.1\) yielded limited progress, and \(w_{\text{hf}} = 0.4\) produced moderate but less consistent gains. These findings suggest that stronger weighting of human feedback can accelerate learning and improve final performance, though optimal values may depend on task complexity and decoder reliability.

\section{CONCLUSION}

This paper introduced a novel framework for RLIHF, leveraging EEG signals to guide robotic policy learning in the scarce of dense reward supervision. By decoding ErrPs as a continuous evaluative signal, the proposed method enables adaptation of agent behavior without requiring explicit human interventions. 
Empirical results from a simulated pick-and-place task using a Kinova Gen2 robotic arm demonstrate that RLIHF agents consistently outperform sparse-reward baselines and achieve performance comparable to agents trained with fully engineered dense rewards. Importantly, these gains were observed even when decoder accuracy varied across subjects, underscoring the robustness of the approach to individual differences in neural signal quality. Further analysis revealed that feedback weighting plays a critical role in balancing exploration and feedback exploitation. Increasing the weight of neural feedback generally accelerated both early and late-phase performance, with the highest weight yielding the strongest overall returns. Future work may explore adaptive weighting schedules for feedback integration and real-world deployment with closed-loop EEG acquisition. In doing so, this line of research opens the door to seamless human-robot collaboration via cognitively unobtrusive neural interfaces, contributing to both assistive system design and the broader advancement of BCI technologies.

\addtolength{\textheight}{-12cm}   

\bibliographystyle{IEEEtran}
\bibliography{reference}

\end{document}